%% file: infotheca_guide.tex
\documentclass[a5paper]{llncs}
\usepackage{tabularx}
\usepackage{amssymb}
\input{settings/settings} 

\bibliography{refs} 

\begin{document}
\label{firstpage}
\sloppy 
\input{settings/language_srp} 

\title{Нови језички модели за српски језик} 

\udk{} 

\begin{main} 

    \begin{abstractenv}  
У раду ће укратко бити приказан историјат развоја језичких модела за српски језик који су засновани на трансформерској архитектури. Биће, такође, представљено неколико нових модела за генерисање и векторизацију текста, обучених на ресурсима Друштва за језичке ресурсе и технологије. Десет одабраних модела за векторизацију српског језика, међу којима су и  два нова модела, биће упоређена на четири задатка обраде природног језика. Биће анализирано који модели су најбољи за изабране задатке, како величина модела и величина скупа за обучавање утичу на њихове перформансе на тим задацима и шта је потребно за обучавање најбољих модела за српски језик.

	\keywords{језички модели, српски језик, векторизација, обрада природног језика.} 
		
	\dates{0.4}{27. јануар 2024.}{24. фебруар 2024.} 
	\end{abstractenv}

& 

    \begin{authorsenv}
	\author{Михаило Шкорић} 
	\email{mihailo.skoric@rgf.bg.ac.rs}  
 	\orcid{0000-0003-4811-8692}  
	\institution{Универзитет у Београду} 
	\institution{Рударско-геолошки факултет} 	
	\institution{Београд, Србија} 	

    \end{authorsenv}

\end{main}

\section{Увод}\label{sec:1}

Почетком двадесет и првог века, дошло је најпре до наглог пораста количине доступних текстуалних података, а потом и до наглог раста рачунарске моћи, што је покренуло талас истраживања заснованих на идеји дубоког учења (\textit{deep learning})~\autocite{lecun2015deep}. У случају обраде природних језика, истраживања кулминирају појавом архитектуре трансформера~\autocite{vaswani2017attention}, која се базира на употреби енкодера, чија је главна намена анализа текста, и декодера, који су задужени за синтезу текста. Први изразито популаран модел овог типа био је \textit{BERT}\footnote{\textit{Bidirectional Encoder Representations from Transformers} -- двосмерно кодирање репрезентације из трансформера}~\autocite{devlin2018bert}, заснован искључиво на трансформерском енкодеру. Овај модел је направио велики помак у обради природних језика, пре свега на задацима који се заснивају на векторизацији текста. Његове варијације, \textit{RoBERTa}\footnote{\textit{Robustly Optimized BERT} -- Робусно оптимизовани \textit{BERT}}~\autocite{liu2019roberta} и \textit{DeBERTa}\footnote{\textit{Decoding-Enhanced BERT} -- \textit{BERT} проширен декодирањем}~\autocite{he2020deberta} и данас постижу најбоље резултате на задацима угњежђивања речи (\textit{word embedding}), анотације речи (нпр. обележавање врсте речи и препознавање именованих ентитета) и класификације реченица и докумената. Са друге стране, појављивање модела \textit{GPT} (генеративни предобучени трансформер)~\autocite{radford2018improving} и \textit{GPT-2}~\autocite{radford2019language} је популаризовало језичке моделе засноване на траснформерском декодеру, а ова група модела  се данас најбрже развија. Модели који комбинују употребу енкодера и декодера, као што су, на пример \textit{BART}~\autocite{lewis2020bart} и \textit{T5}~\autocite{raffel2020exploring}, остају недовољно запажени упркос изванредним резултатима које постижу на задацима трансформације текста, као што су машинско превођење, сумaризација и прилагођавање стила.

\subsection{Преглед објављених модела за српски језик}
Језички модели засновани на архитектури трансформера су направили продор у српски језик путем вишејезичних модела, најпре кроз \textit{MBERT}\footnote{\textit{Multilingual BERT} -- вишејезични \textit{BERT}}~\autocite{devlin2018bert}, а потом и кроз \textit{XLM-RoBERTa} модел\footnote{\textit{Cross-lingual Language Model} -- Међујезички језички модел}~\autocite{conneau2019unsupervised}, за чије обучавање је коришћено око 4 милијарде токена из тесктова писаних на српском или другом блиско-сродном језику (хрватски, босански). Потоњи модел објављен је у децембру 2019. године у две варијанте, \textit{base} (279 милиона параметара) и \textit{large} (561 милион параметара). И данас се, као један од највећих модела за векторизацију, употребљава у обради српског језика и притом остварује добре резултате, поготово након дообучавања.


Почетком 2021. године, на платформи \textit{Huggingface}\footnote{\href{https://huggingface.co}{Huggingface}, највеће веб чвориште за објављивање језичких модела.} објављен је  модел под називом \textit{БЕРТић} (\textit{classla/bcms-bertic})~\autocite{ljubevsic2021berti}, базиран на архитектури \textit{ELECTRA}~\autocite{clark2020electra}, са 110 милиона параметара, обучаван на корпусу од преко 8 милијарди токена, босанског (800 милиона), хрватског (5.5 милијарди), црногорског (80 милиона) и српског језика (2 милијарде).


Касније те исте године, направљени су и објављени први специфични модели за српски језик у оквиру једног ширег језичког истраживања за македонски језик~\autocite{dobreva2022macedonizer}. Прецизније, објављена је српска верзија \textit{RoBERTa-base} модела, \textit{macedonizer/sr-roberta-base} (120 милиона параметара) и српска верзија \textit{GPT2-small} модела, \textit{macedonizer/sr-gpt2} (130 милиона параметара). Оба модела су обучавана на корпусу српске Википедије и подржавају само ћирилично писмо.


Недуго потом предузет је сличан подухват, при којем је обучено пет \textit{RoBERTa-base} модела за српски језик~\autocite{cvejic2022prepoznavanje}. Иницијални модел, \textit{Andrija/SRoBERTa}, имао је 120 милиона параметара и обучаван је на малом корпусу од 18 милиона токена познатом под именом \textit{Leipzig}~\autocite{biemann2007leipzig}, док су потоња четири модела имала по 80 милиона параметара, при чему је сваки обучаван на све већем корпусу. За модел \textit{Andrija/SRoBERTa-base} додат је корпус \textit{OSCAR}~\autocite{suarez2019asynchronous} (220 милиона токена), за модел \textit{Andrija/SRoBERTa-L} је поред њега додат и \textit{srWAc}~\autocite{ljubevsic2014bs} (490 милиона токена), за модел \textit{Andrija/SRoBERTa-XL} је уз претходне додат и део корпуса \textit{cc100-hr} (21 милијарда токена) и \textit{cc100-sr} (5.5 милијарди токена)~\autocite{wenzek-etal-2020-ccnet}, док су за модел \textit{Andrija/SRoBERTa-F} сви поменути корпуси коришћени у целости.


Крајем 2022. године објављена су три експериментална генеративна модела за српски језик~\autocite{vskoric2023композитне}. Контролни модел \textit{procesaur/gpt2-srlat}  је био поново заснован на \textit{GPT2-small} архитектури, имао је 138 милиона параметара и био је обучен на исечку корпуса Друштва за језичке ресурсе и технологије (260 милиона токена)~\autocite{CvRS2023LRS}. Друга два модела, \textit{procesaur/gpt2-srlat-sem} и \textit{procesaur/gpt2-srlat-synt}, настала су дообучавањем контролног модела коришћењем два специјално припремљена корпуса са циљем засебног моделовања семантике, односно, синтаксе текста. Три модела су потом употребљена за експеримент комбиновања језичких модела на задатку класификације реченица~\autocite{vskoric2023transformer}.


Почетком наредне године, истраживачи са Универзитета у Нишу објавили су модел \textit{JelenaTosic/SRBerta} (75 милиона параметара) који је такође заснован на  \textit{RoBERTa-base} архитектури, а који је обучаван помоћу корпуса \textit{OSCAR}~\autocite{suarez2019asynchronous}.Занимљиво је да је овај модел, као и његова друга верзија (\textit{nemanjaPetrovic/SRBerta}, 120 милиона параметара), пре објављивања дообучен над текстовима из домена права~\autocite{bogdanovicsrberta}. У периоду између објављивања ова два модела, објављен је и \textit{aleksahet/xlm-r-squad-sr-lat}~\autocite{cvetanovic2023synthetic}, први модел за одговарање на питања на српском језику, настао прилагођавањем \textit{XLM-RoBERTа} модела помоћу скупа података \textit{SQuAD}~\autocite{rajpurkar2018know}, преведеног на српски језик.


Средином 2023. године објављена су још два генеративна модела заснована на ГПТ архитектури. Оба модела су обучавана над истим скупом података: над корпусима Друштва за језичке ресурсе и технологије~\autocite{CvRS2023LRS}, докторским дисертацијама преузетим са платформе НАРДУС,\footnote{\href{https://nardus.mpn.gov.rs/}{НАРДУС} -- Национални репозиторијум докторских дисертација са свих универзитета у Србији.} корпусу јавног дискурса српског језика Института за српски језик САНУ под називом \textit{PDRS}~\autocite{11356/1752} и додатним јавно доступним корпусима са веба, као што су већ поменути \textit{srWAc}~\autocite{ljubevsic2014bs} и \textit{cc100-sr}~\autocite{wenzek-etal-2020-ccnet}. Укупан број токена у овом скупу података броји око 4 милијарде токена. Већи модел, \textit{jerteh/gpt2-orao},\footnote{\href{https://huggingface.co/jerteh/gpt2-orao}{jerteh/gpt2-orao}} броји 800 милиона параметара, заснован је на архитектури \textit{GPT2-large} и представља тренутно највећи доступни модел предобучен за српски језик. Мањи модел, \textit{jerteh/gpt2-vrabac},\footnote{\href{https://huggingface.co/jerteh/gpt2-vrabac}{jerteh/gpt2-vrabac}} броји 136 милиона параметара и заснован је на архитектури \textit{GPT2-small}. Оба модела су обучавана коришћењем рачунарских ресурса Националне платформе за вештачку интелигенцију Србије. Осим корпуса за обучавање, ова два модела деле и речник токена и токенизатор, специјално опремљен да упарује ћирилична и латинична слова, омогућујући њихову равноправну подршку.


Након објављивања генеративног модела од 800 милиона (\textit{jerteh/gpt2-orao}) параметара, фокус се полако помера на дообучавање великих модела коришћењем текстова на српском језику. Тако су објављена два модела заснована на архитектури \textit{Alpaca}~\autocite{taori2023alpaca}, \textit{datatab/alpaca-serbian-3b-base} (3 милијарде параметара) и \textit{datatab/alpaca-serbian-7b-base} (7 милијарди параметара), a најављено је и објављивање још једног модела исте величине, заснованог на архитектури \textit{Mistral-7b}~\autocite{jiang2023mistral}, који је обучаван на хрватским, босанским и српским текстовима који броје 11,5 милијарди токена. Исти корпус од 11,5 милијарди токена коришћен је и за дообучавање \textit{XLM-RoBERTa-large} модела у циљу поређења перформанси дообучаваних модела у односу на моделе који су обучавани од почетка (\textit{од нуле}). Нови модел је објављен под именом \textit{classla/xlm-r-bertic} и броји 561 милион параметара, колико има и оригинални \textit{XLM} модел.


Коначно, на скупу података над којим су обучавани \textit{jerteh/gpt2-orao} и \textit{jerteh/gpt2-vrabac}, обучена су још два модела за векторизацију текста. Већи модел, \textit{jerteh/Jerteh-355}\footnote{\href{https://huggingface.co/jerteh/jerteh-355}{jerteh/Јerteh-355}}, заснован је на \textit{RoBERTa-large} архитектури и броји 355 милиона параметара, док је мањи модел, \textit{jerteh/Jerteh-81},\footnote{\href{https://huggingface.co/jerteh/jerteh-81}{jerteh/Јerteh-81}} заснован на \textit{RoBERTa-base} архитектури и броји 81 милион параметара. Као и код модела \textit{jerteh/gpt2-orao}, циљ је био да се модели обуче на што квалитетнијем корпусу текстова.  У овом раду биће представљено испитивање перформанси ова два модела и њихово поређење са перформансама других одабраних модела како би се установило њихово место у хијерархији језичких модела за векторизацију текста на српском језику.


\subsection{Поставка експеримента}
У претходном одељку је указано на постојање већег броја вишејезичних модела који, у мањој или већој мери, подржавају обраду српског језика, као и на двадесетак модела који су припремљени специјално за обраду српског језика. Објављени модели се међусобно разликују према неколико особина: породици (архитектури) модела и броју његових параметара, речнику, односно токенизатору, на којем се заснивају, скупу који је коришћен за њихово обучавање, задатку на којем је модел обучаван и дужини обучавања. Треба напоменути да неке од ових информација недостају за неке од модела, али и да су неке информације које су доступне (пре свега особине скупа који је коришћен за обучавање) непроверљиве.


У наставку, рад ће се фокусирати на десет одабраних модела за векторизацију (општег типа). Основне информације о тим моделима биће представљене у одељку~\ref{sec:2}, експеримент поређења њихових перформанси на четири припремљена задатка бити приказан у  одељку~\ref{sec:3}, а резултати експеримената ће бити приказани и размотрени у одељку~\ref{sec:4} Напослетку, у одељку~\ref{sec:5}, биће предложен процес обучавања нових модела за српски језик. Овај рад се неће фокусирати на генеративне моделе услед недостатка поузданог (али аутоматског) механизма за мерење њихових перформанси. Још увек није објављен ниједан енкодер-декодер модел специјално развијен за српски језик.


\section{Одабрани модели за векторизацију текста}
\label{sec:2}

За потребе овог рада, од претходно поменутих модела (одељак~\ref{sec:1}) одабрано је десет који ће бити детаљније анализирани. У тих десет улазе најпре четири \textit{SRoBERTa} модела, који су, услед тога што се разликују искључиво по скупу података за обучавање, врло погодни за овај експеримент. Даље, ту су најстарији модел, \textit{classla/bcms-bertic} и најновији модел, \textit{classla/xlm-r-bertic}, које је објавио центар за јужнословенске језике \textit{CLASSLA}, као и два најпопуланрија вишејезична модела \textit{xlm-roberta-base} и \textit{xlm-roberta-large}. Коначно, ту су два модела која се први пут представљају у овом раду, модели \textit{jerteh-81} и \textit{jerteh-355}, који су обучени над ресурсима Друштва за језичке ресурсе и технологије.


Основне карактеристике ових десет модела приказане су у Табели~\ref{table:1}.

\begin{table}
\newcolumntype{Y}{>{\centering\arraybackslash}X}
\centering
\renewcommand{\arraystretch}{2}
\begin{tabularx}{\textwidth}{ | c | Y | Y | Y | Y | c | c | Y | Y | Y | Y | }

  \hline 
  редни број & 1 & 2& 3 & 4& 5 & 6& 7 & 8& 9 & 10 \\ \hline  \hline
 Идентификатор & {\rotatebox[origin=c]{90}{ Andrija/SRoBERTa-base }}  &   {\rotatebox[origin=c]{90}{Andrija/SRoBERTa-L}}  &   {\rotatebox[origin=c]{90}{Andrija/SRoBERTa-XL}}  &   {\rotatebox[origin=c]{90}{Andrija/SRoBERTa-F}} &  {\rotatebox[origin=c]{90}{classla/bcms-bertic}} &  {\rotatebox[origin=c]{90}{classla/xlm-r-bertic}} &  {\rotatebox[origin=c]{90}{xlm-roberta-base}} &  {\rotatebox[origin=c]{90}{xlm-roberta-large}} &  {\rotatebox[origin=c]{90}{jerteh/jerteh-81}} &  {\rotatebox[origin=c]{90}{jerteh/jerteh-355}}  \\ \hline  \hline

Речник токена & \multicolumn{4}{ c |}{SRoBERTa} & bertic &  \multicolumn{3}{ c }{XLM-R} &  \multicolumn{2}{| c |}{jerteh}\\ \hline 
Архитектура & \multicolumn{4}{ c |}{RoBERTa} & ELE. &  \multicolumn{3}{ c }{XLM-R} &  \multicolumn{2}{| c |}{RoBERTa}\\ \hline 
Величина модела & \multicolumn{4}{ c |}{80}& 110 & 561 & 279 & 561 & 81 & 355 \\ \hline  
Величина скупа & 500 & 1000 & 3750 & 5700 & 8400 & 11500 & \multicolumn{2}{ c |}{ 4000*} &  \multicolumn{2}{ c |}{ 4000}\\ \hline  \hline  
Српски  & \checkmark  & \checkmark  & \checkmark  & \checkmark & \checkmark & \checkmark & \checkmark & \checkmark & \checkmark & \checkmark \\ \hline  
Хрватски  &   &   & \checkmark  & \checkmark & \checkmark & \checkmark & \checkmark & \checkmark &  &  \\ \hline  
Босански  &   &   &   &  & \checkmark & \checkmark & \checkmark & \checkmark &  &  \\ \hline  
Црногорски  &   &   &   &  & \checkmark &  & \checkmark & \checkmark &  &  \\ \hline

\end{tabularx}
\vspace{0.5cm}
\caption{Десет одабраних модела за векторизацију текста на српском језику и њихове особине: речник токена на којем су засновани, архитектура модела, величина модела изражена у милионима параметара и величина скупа изражена у милионима токена. Подаци су преузети са платформе \textit{HuggingFace}.
*Величина скупа за обучавање код модела 7 и 8 (\textit{xlm-roberta-base}, \textit{xlm-roberta-large}) односи се на део скупа на српском, хрватском или другом сродном језику. Доњи део табеле приказује на којем од ових језика су модели обучавани.} 
\label{table:1}
\end{table}

У приложеној табели, као и из описа модела у претходном одељку, види се да је најпопуларнија архитектура \textit{RoBERTa} (6 од 10 одабраних модела), a додатна три модела заснована су на блиској, \textit{XLM-RoBERTa} архитектури. Преостали модел, \textit{bcms-bertic}, заснива се на ELECTRA архитектури и једини је од одабраних који није обучаван на задатку моделовања маскираног језика (предвиђања делова текста маскираних иза неке специјалне етикете).


Величина одабраних модела варира од 80 (за четири \textit{SRoBERTa} модела) па до преко 560 милиона параметара (за моделе засноване на \textit{XLM-RoBERTa-large}). Величина скупа за обучавање варира од 500 милиона за модел 1 (\textit{SRoBERTa-base}), па до чак 11,5 милијарди токена за модел 6 (\textit{classla/xlm-r-bertic}), при чему треба напоменути да он није обучаван од нуле, већ је у питању \textit{xlm-roberta-large} модел дообучен за хрватски, босански и српски језик. Само четири од десет модела су обучавана искључиво над корпусом српских текстова. У питању су модели 1, 2, 9 и 10, тј. прва два \textit{SRoBERTa} модела, \textit{jerteh/jerteh-81} и \textit{jerteh/jerteh-355}.


Битно је напоменути и да десет приказаних модела користе само четири различита речника токена, односно токенизатора:
\begin{description}
\item[$X_1$] \textit{SRoBERTa} токенизатор - прва 4 модела;
\item[$X_2$] \textit{bertic} токенизатор - модел број 5;
\item[$X_3$] \textit{XLM-R} токенизатор -  модели 6 до 8;
\item[$X_4$] \textit{jerteh} токенизатор - последња 2 модела (9 и 10).
\end{description}

\section{Поставка евалуацијe перформанси модела}
\label{sec:3}

 Десет одабраних модела је евалуирано на четири засебна задатка како би се упоредиле њихове перформансе:


\begin{description}
    \item[$T_1$] Моделовање маскираног језика (погађање недостајућих токена);
    \item[$T_2$] Израчунавање (семантичке) сличности између реченица;
    \item[$T_3$] Обележавање врстом речи;
    \item[$T_4$] Препознавање именованих ентитета.
\end{description}
Прва два задатка припадају групи такозваних узводних задатака (\textit{upstream}), то јест задатака који користе моделе у њиховом основном стању, док друга два задатка припадају групи низводних задатака (\textit{downstream}) јер захтевају да се модели фино подесе (\textit{fine-tune}) и тестирају на специјално припремљеном скупу података.


\subsection{Евалуација модела на узводним задацима}
Као што је већ поменуто, за узводне задатке није неопходно прилагођавање модела, тако да је потребно само де се припреме скупове за тестирање.


Како би се спровела евалуација модела на задатку моделовања маскираног језика ($T_1$) најпре је припремљен специјалан скуп података у виду текстова у којима су у свакој реченици по један насумично изабрани токен маскирани -- по један токен је сакривен, на пример иза маске \texttt{<MASK>}. За текстуалну грађу су коришћена четири извора:


\begin{description}
    \item[$Y_1$] \textit{Дечко}, српски превод романа \textit{Подросток} од Достојевског;
    \item[$Y_2$] \textit{Младић}, алтернативни  превод \textit{Дечка};
    \item[$Y_3$] \textit{Пут око света за 80 дана}, српски превод романа Жила Верна;
    \item[$Y_4$] \textit{Пут око свијета у 80 дана}, хрватски превод романа Жила Верна.
\end{description}

Прва два извора нису коришћена за обучавање ниједног модела, док су друга два већ дуго доступни на вебу~\autocite{vitas2008tour} и стога су вероватно коришћени за обучавање већине, ако не и свих, наведених модела.
Како неки модел не би био у посебној предности, текстови су токенизовани коришћењем сва четири токенизатора ($X_1$ до $X_4$), потом маскирани, а онда је пред сваким од модела био задатак да одмаскира свих шеснаест припремљених текстова (четири извора токенизована и маскирана на четири начина). У свакој реченици био је маскиран по један токен, а модели су током евалуације за његово место нудили по три кандидата. За процену резултата теста узимана је мера тачности на овом задатку, при чему се као погодак рачунало свако одмаскирање код кога је маскирани, то јест тражени, токен био у скупу кандидата које је модел понудио за задату реченицу.


За потребе евалуације на задатку израчунавања сличности између реченица ($T_2$), коришћене су реченице из истих романа, то јест два пара паралелизованих романа ($Y_1$  и $Y_2$, односно $Y_3$ и $Y_4$), али припремљене као триплети. С обзиром на то да су романи претходно паралелизовани на нивоу реченице, било је лако направити парове реченица које имају исто значење. Сваки триплет је употпуњавала додатна реченица из романа парњака, која дели што је више могуће токена са контролном реченицом и има сличну дужину, али је повучена из неког другог места у тексту. Пример триплета:


\begin{enumerate}
\item \underline{"}Zaista, ko \underline{ne bi} obišao svet i za manju cenu\underline{?"} (контролна реченица, $Y_3$: \textit{Пут око света за 80 дана})
    \item "Doista, nije li i za manje od toga vrijedno izvršiti put oko svijeta?" (парњак, $Y_4$: \textit{Пут око свијета у 80 дана})
    \item \underline{"}He! he! pa konačno zašto \underline{ne bi} uspio\underline{?"} (лажни парњак, $Y_4$: \textit{Пут око свијета у 80 дана})
\end{enumerate}

Задатак модела је био да у задатим триплетима препознају правог парњака (сличност између прве и друге реченице треба да буде већа него између прве и треће), а за процену резултата теста је узимана тачност на том задатку. Да би се израчунао реченични вектор, модел најпре додели векторске вредности сваком токену у реченици, а потом се вредност тих вектора усредњава како би се добила векторска репрезентација реченице. Сличност између две реченице се израчунава као разлика броја 1 и косинусне удаљености израчунатих реченичних вектора.


\subsection{Евалуација модела на низводним задацима}
За потребе евалуације модела преостала два предвиђена задатка, модели су дообучени и тестирани на посебним скуповима података. За обележавање врстом речи ($T_3$) коришћен је јавно доступни скуп \textit{SrpKor4Tagging}~\autocite{stankovic2020machine} (триста педесет хиљада обележених токена), док је за препознавање именованих ентитета ($T_4$) коришћен други јавно доступни скуп \textit{SrpELTeC-gold}~\autocite{todorovic2021serbian}. У оба случаја, модели су дообучени на $90\%$ обележених реченица из сваког скупа и тестирани на преосталих $10\%$. Како се ради о проблему вишекласне класификације, за процену резултата ова два задатка узете су $F_1$-мере остварене приликом класификације над реченицама из скупа за тестирање.


\section{Резултати евалуације}\label{sec:4}

Резултати  првог теста ($T_1$), то јест просечна тачност одабраних модела на задатку допуњавања недостајућег токена у сваком од шеснаест припремљених текстова, приказани су у табели~\ref{table:2}.


\begin{table}[ht!]
\newcolumntype{Y}{>{\centering\arraybackslash}X}
\renewcommand{\arraystretch}{1.1}
\centering
\vspace{0.5cm}
\begin{tabularx}{\textwidth}{ | c | Y | Y | Y | Y | Y | Y | Y | Y | Y | Y |}
  \hline   редни број & 1 & 2& 3 & 4& 5 & 6& 7 & 8& 9 & 10 \\ \hline  \hline  
  & {\rotatebox[origin=c]{90}{ Andrija/SRoBERTa-base }}  &   {\rotatebox[origin=c]{90}{Andrija/SRoBERTa-L}}  &   {\rotatebox[origin=c]{90}{Andrija/SRoBERTa-XL}}  &   {\rotatebox[origin=c]{90}{Andrija/SRoBERTa-F}}
  &  {\rotatebox[origin=c]{90}{classla/bcms-bertic}}
  &  {\rotatebox[origin=c]{90}{classla/xlm-r-bertic}} &  {\rotatebox[origin=c]{90}{xlm-roberta-base}} &  {\rotatebox[origin=c]{90}{xlm-roberta-large}} &  {\rotatebox[origin=c]{90}{jerteh/jerteh-81}} &  {\rotatebox[origin=c]{90}{jerteh/jerteh-355}}  \\ \hline  \hline
$X_1$-$Y_1$ & 0.43 & 0.63 & 0.66 & 0.70 & / & 0.43 & 0.46 & 0.51 & 0.70 & \textbf{0.75}
\\ \hline
$X_1$-$Y_2$ & 0.43 & 0.62 & 0.64 & 0.69 & / & 0.42 & 0.46 & 0.50 & 0.69 & \textbf{0.73}\\ \hline
$X_1$-$Y_3$ & 0.37 & 0.56 & 0.59 & 0.63 & / & 0.34 & 0.38 & 0.43 & 0.66 & \textbf{0.72}\\ \hline
$X_1$-$Y_4$ & 0.36 & 0.55 & 0.64 & \textbf{0.68} & / & 0.34 & 0.38 & 0.42 & 0.58 & 0.63\\ \hline
$X_2$-$Y_1$ & 0.36 & 0.47 & 0.51 & 0.54 & / & 0.47 & 0.50 & 0.55 & 0.57 & \textbf{0.60}\\ \hline
$X_2$-$Y_2$ & 0.37 & 0.48 & 0.51 & 0.54 & / & 0.46 & 0.50 & 0.54 & 0.56 & \textbf{0.59}\\ \hline
$X_2$-$Y_3$ & 0.31 & 0.41 & 0.45 & 0.48 & / & 0.42 & 0.45 & 0.50 & 0.50 & \textbf{0.54}\\ \hline
$X_2$-$Y_4$ & 0.31 & 0.42 & 0.47 & \textbf{0.51} & / & 0.42 & 0.46 & \textbf{0.50} & 0.47 & \textbf{0.51}\\ \hline
$X_3$-$Y_1$ & 0.37 & 0.49 & 0.52 & 0.54 & / & 0.48 & 0.50 & 0.55 & 0.57 & \textbf{0.60}\\ \hline
$X_3$-$Y_2$ & 0.37 & 0.48 & 0.51 & 0.54 & / & 0.46 & 0.50 & 0.54 & 0.57 & \textbf{0.59}\\ \hline
$X_3$-$Y_3$ & 0.30 & 0.41 & 0.44 & 0.47 & / & 0.41 & 0.45 & 0.49 & 0.50 & \textbf{0.54}\\ \hline
$X_3$-$Y_4$ & 0.31 & 0.42 & 0.47 & \textbf{0.51} & / & 0.41 & 0.46 & \textbf{0.50} & 0.47 & \textbf{0.50}\\ \hline
$X_4$-$Y_1$ & 0.42 & 0.60 & 0.63 & 0.67 & / & 0.43 & 0.47 & 0.51 & 0.73 & \textbf{0.78}\\ \hline
$X_4$-$Y_2$ & 0.41 & 0.58 & 0.61 & 0.65 & / & 0.41 & 0.45 & 0.49 & 0.71 & \textbf{0.75}\\ \hline
$X_4$-$Y_3$ & 0.35 & 0.53 & 0.55 & 0.60 & / & 0.33 & 0.38 & 0.42 & 0.69 & \textbf{0.76}\\ \hline
$X_4$-$Y_4$ & 0.34 & 0.50 & 0.58 & 0.62 & / & 0.33 & 0.37 & 0.41 & 0.62 & \textbf{0.66}\\ \hline
просек & 0.36 & 0.51 & 0.55 & 0.59 & / & 0.41 & 0.45 & 0.49 & 0.60 & \textbf{0.64}\\ \hline

\end{tabularx} 
\vspace{0.2cm}
\caption{Тачност модела у погађању токена (из три покушаја) на задатку моделовања маскираног језика над шеснаест припремљених маскираних текстова и њихова просечна тачност. Текстови су обележени (на почетку сваког реда) јединственим комбинацијама ознака токенизатора $X$ и извора $Y$. У сваком реду најбољи резултат ($\pm1\%$) обележен је подебљањем.}
\label{table:2}
\end{table}


У приложеним резултатима је може се уочити супериорност новог модела, \textit{jerteh-355}, који остварује бољи резултат од осталих модела у тринаест од шеснаест случајева, односно бољи или исти резултат ($\pm1\%$) у петнаест од шеснаест случајева. Осим тога, у девет од дванаест случајева модел \textit{jerteh-355} надмашује друге моделе чак и када обрађује текст маскиран токенизатором тих модела. Једини модел који успева да га надмаши у два случаја је \textit{SRoBERTa-F}, који се најчешће показује као најбољи у обради извора ($Y_4$) писаног на хрватском језику, који је у великом проценту био укључен у његов скуп за обучавање. Ипак, у просеку, његова тачност на овом тесту је нижа и од оне коју остварује други нови модел, \textit{jerteh-81}. Модел 5 (\textit{classla/bcms-bertic}), који, за разлику од осталих, није обучаван на задатку моделовања маскираног језика, није био укључен у евалуацију на овом задатку, јер би био у неповољном положају. 


Резултати теста израчунавања сличности између реченица ($T_2$) приказани су у табели~\ref{table:3}. Вредности приказују тачност модела при препознавању реченица са истим/сличним значењем у триплетима екстрахованим из два српска превода истог романа, $Y_1$ i $Y_2$ (први ред вредности), из српског и хрватског превода истог романа, $Y_3$ и $Y_4$ (други ред), и просечну тачност (трећи ред).


\begin{table}
\newcolumntype{Y}{>{\centering\arraybackslash}X}
\renewcommand{\arraystretch}{1.1}
\centering
\begin{tabularx}{\textwidth}{ | c | Y | Y | Y | Y | Y | Y | Y | Y | Y | Y | }

  \hline   редни број & 1 & 2& 3 & 4& 5 & 6& 7 & 8& 9 & 10 \\ \hline  \hline  
  & {\rotatebox[origin=c]{90}{ Andrija/SRoBERTa-base }}  &   {\rotatebox[origin=c]{90}{Andrija/SRoBERTa-L}}  &   {\rotatebox[origin=c]{90}{Andrija/SRoBERTa-XL}}  &   {\rotatebox[origin=c]{90}{Andrija/SRoBERTa-F}} &  {\rotatebox[origin=c]{90}{classla/bcms-bertic}} &  {\rotatebox[origin=c]{90}{classla/xlm-r-bertic}} &  {\rotatebox[origin=c]{90}{xlm-roberta-base}} &  {\rotatebox[origin=c]{90}{xlm-roberta-large}} &  {\rotatebox[origin=c]{90}{jerteh/jerteh-81}} &  {\rotatebox[origin=c]{90}{jerteh/jerteh-355}}  \\ \hline  \hline
$Y_1$-$Y_2$ & 0.93 & \textbf{0.95} & \textbf{0.96} & \textbf{0.96} & 0.92 & 0.76 & 0.90 & 0.87 & \textbf{0.95} & \textbf{0.95}\\ \hline
$Y_3$-$Y_4$ & 0.83 & 0.89 & \textbf{0.92} & \textbf{0.91} & 0.79 & 0.66 & 0.78 & 0.71 & 0.89 & 0.83\\ \hline
просек & 0.88 & 0.92 & \textbf{0.94} & \textbf{0.93} & 0.85 & 0.71 & 0.84 & 0.79 & 0.92 & 0.89\\ \hline 

\end{tabularx} 
\vspace{0.3cm}
\caption{Резултати одабраних модела на задатку препознавања реченица са истим значењем у триплетима екстрахованим из превода Достојевског ($Y_1$-$Y_2$), Верна ($Y_3$-$Y_4$), као и у просеку. У сваком реду најбољи резултат ($\pm1\%$) обележен је подебљањем.}
\label{table:3}
\end{table}

Резултати за први низ триплета су веома добри за неколико модела: \textit{SRoBERTa-L}, \textit{SRoBERTa-XL}, \textit{SRoBERTa-F}, \textit{jerteh-81} и \textit{jerteh-355} остварују сличну тачност од преко 95\%. Модел \textit{SRoBERTa-XL} остварује најбоље резултате, уз малу маргину, али и најбољи резултат за други низ триплета који садржи реченице на хрватском језику (92\% тачности), па самим тим  има и најбољи просечни резултат на овом задатку. Једини други модел који остварује тачност од преко 90\% за други низ триплета је \textit{SRoBERTa-F}, што је и очекивано јер су ова два модела обучавана на хрватским текстовима.


Резултати који су модели остварили на низводним задацима $T_3$ (обележавање врсте речи) и $T_4$ (препознавање именованих ентитета) приказани су у виду $F_1$-мера у Табели~\ref{table:4}.

\begin{table}
\newcolumntype{Y}{>{\centering\arraybackslash}X}
\renewcommand{\arraystretch}{1.1}
\centering
\begin{tabularx}{\textwidth}{ | c | Y | Y | Y | Y | Y | Y | Y | Y | Y | Y | }

  \hline р.б. & 1 & 2& 3 & 4& 5 & 6& 7 & 8& 9 & 10 \\ \hline  \hline  
  & {\rotatebox[origin=c]{90}{ Andrija/SRoBERTa-base }}  &   {\rotatebox[origin=c]{90}{Andrija/SRoBERTa-L}}  &   {\rotatebox[origin=c]{90}{Andrija/SRoBERTa-XL}}  &   {\rotatebox[origin=c]{90}{Andrija/SRoBERTa-F}} &  {\rotatebox[origin=c]{90}{classla/bcms-bertic}} &  {\rotatebox[origin=c]{90}{classla/xlm-r-bertic}} &  {\rotatebox[origin=c]{90}{xlm-roberta-base}} &  {\rotatebox[origin=c]{90}{xlm-roberta-large}} &  {\rotatebox[origin=c]{90}{jerteh/jerteh-81}} &  {\rotatebox[origin=c]{90}{jerteh/jerteh-355}}  \\ \hline  \hline

$T_3$ & 0.974 & 0.980 & 0.982 & 0.982 & \textbf{0.986} & \textbf{0.987} & 0.984 & \textbf{0.986} & 0.985 & \textbf{0.986}\\ \hline
$T_4$ & 0.908 & 0.922 & 0.929 & 0.935 & \textbf{0.942} & \textbf{0.942} & 0.933 & 0.935 & 0.928 & 0.928\\ \hline

\end{tabularx} 
\vspace{0.5cm}
\caption{$F_1$-мера коју су модели остварили на задацима $T_3$ (обележавање врсте речи) и $T_4$ (препознавање именованих ентитета). У сваком реду најбољи резултат ($\pm0.1\%$)  обележен је подебљањем.}
\label{table:4}
\end{table}

Из резултата приказаних у табели~\ref{table:4} види се да на задатку $T_3$ (обележавање врстом речи)  девет од десет модела остварује јако добре резултате (преко 98\%), при чему се резултати које остварују четири најбоља модела (\textit{classla/bcms-bertic}, \textit{classla/xlm-r-bertic}, \textit{xlm-roberta-large} и \textit{jerteh/jerteh-355}) разликују за мање од 0.02\%, указујући да се модели полако приближавају горњим границама перформанси када је у питању овај задатак.


Када су у питању резултати остварени на последњем задатку $T_4$ (препознавање именованих ентитета), најбоље перформансе приказали су модели \textit{classla/bcms-bertic} и \textit{classla/xlm-r-bertic}, при чему су разлике у $F_1$-мерама нешто више него на задатку $T_3$ ($\sim4\%$ за задатак $T_4$ у односу на $\sim1\%$ за задатак $T_3$), али и даље знатно ниже него што су разлике на узводним задацима (чак $\sim28\%$ за задатак $T_1$).


У наредном одељку биће разматрани резултати које су модели остварили као и разлози који су довели до тих резултат, са циљем да се установе најповољнији услови за обучавање модела за српски језик у будућности.


\section{Дискусија}\label{sec:5}

Претходно приказани резултати евалуације модела (табеле~\ref{table:2}--\ref{table:4}), показују да не постоји један модел, или група модела, који су најбољи у општем случају, већ су се различити модели показали као бољи (или лошији) на различитим задацима. У наставку, сваки од задатака ће бити посматран појединачно, пре свега у светлу односа остварених перформанси према величини модела, величини скупова за њихово обучавање и квалитету тих скупова.


\subsection{Моделовање маскираног језика}

Резултати евалуације модела на задатку моделовања маскираног језика (табела~\ref{table:2}) показују убедљиву предност модела \textit{jerteh/jerteh-355}, са тим да добре резултате остварује и \textit{jerteh/jerteh-81}, који је други најбољи модел за овај задатак када се посматрају просечне вредности. Како ова два модела користе исти скуп података за обучавање, то указује да би управо овај скуп могао бити разлог добрих резултата.

\begin{figure}[t]
\centering
\includegraphics[scale=0.25]{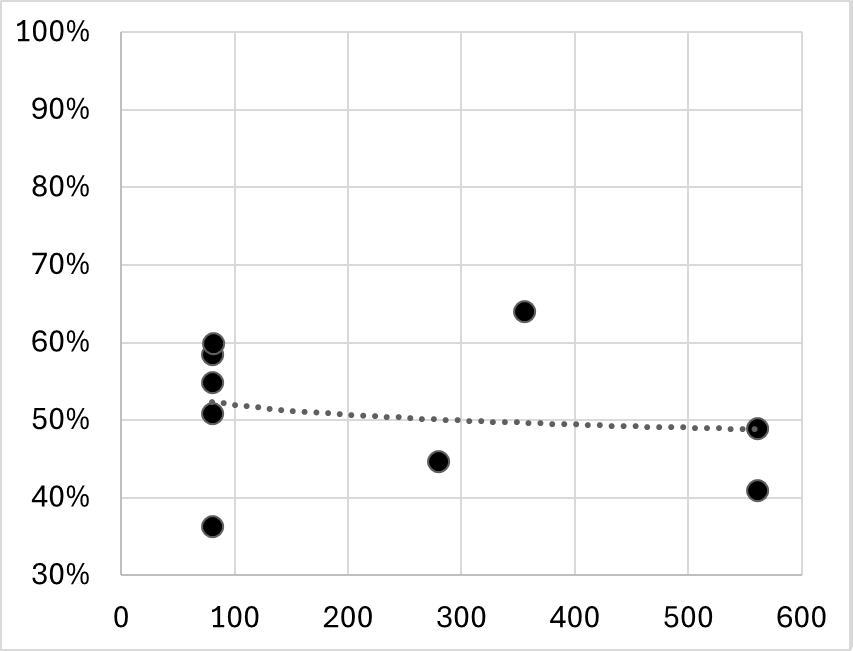}
\includegraphics[scale=0.25]{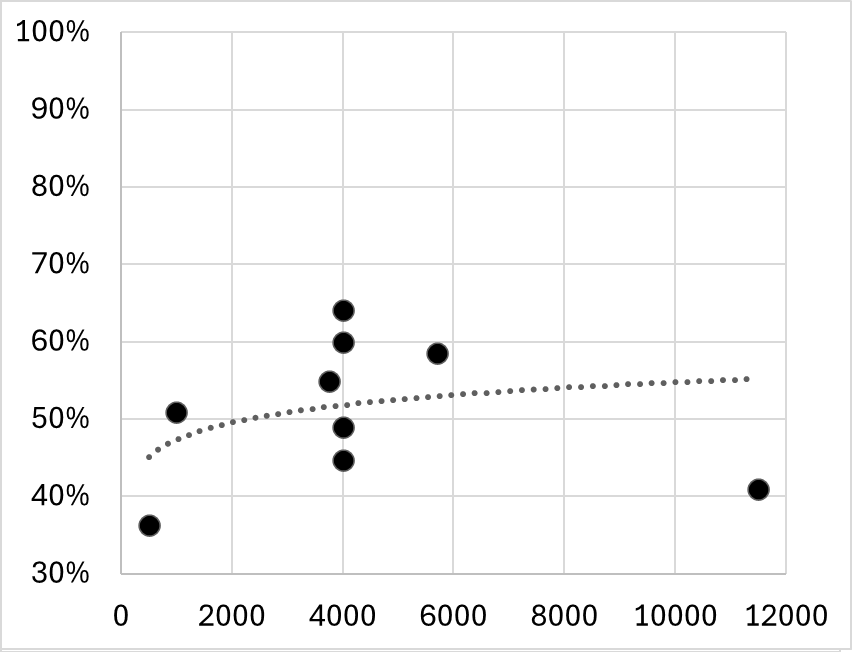}
\caption{Тачност модела на задатку моделовања маскираног језика према величини тих модела (лево), као и према величини скупова за обучавање модела (десно). Приказана крива тренда одговара логаритамској функцији.}
\label{fig:slika1}
\end{figure}
\begin{figure}[ht!]
\centering
\includegraphics[scale=0.25]{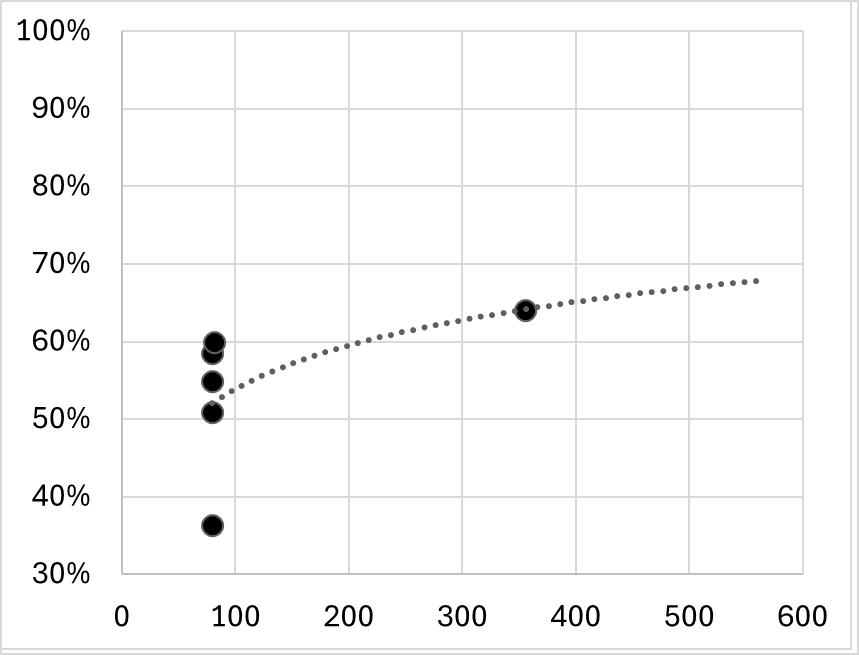}
\includegraphics[scale=0.25]{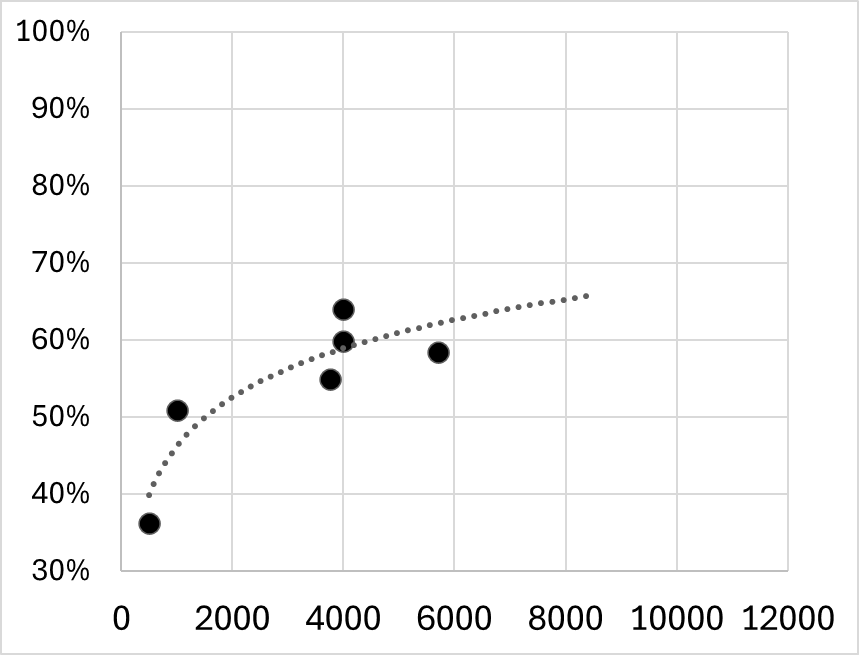}
\caption{Тачност модела на задатку моделовања маскираног језика према величини тих модела (лево), као и према величини скупова за обучавање модела (десно), при чему су уклоњени резултати модела заснованих на \textit{XLM-R} архитектури. Приказана крива тренда одговара логаритамској функцији.}
\label{fig:slika2}
\end{figure}

Просечна тачност модела на задатку $T_1$ према њиховој величини, односно према  величини скупа који је коришћен за њихово обучавање приказан је на слици~\ref{fig:slika1}. На први поглед, приметни су неки истакнути изузеци, пре свега модели засновани на \textit{XLM-R} архитектури, који остварују неке од најлошијих резултата на овом задатку. Уколико се њихови резултати уклоне, појављују се нови трендови (Слика~\ref{fig:slika2}). Дакле, када посматрамо само \textit{RoBERTa} моделе, чини се да већи модел (не баш убедљиво) и већи скуп за његово обучавање (врло убедљиво) повољно утичу на перформансе модела.


\subsection{Израчунавање сличности између реченица}

Приликом евалуације задатка $T_2$ установљено је да најбоље резултате остварују модели \textit{SRoBERTa-L}, \textit{SRoBERTa-XL}, \textit{SRoBERTa-F}, \textit{jerteh-81} и \textit{jerteh-355} када је у питању препознавање сличних реченица на српском језику, а \textit{SRoBERTa-XL} и \textit{SRoBERTa-F} када је у питању препознавање сличних реченица између српског и хрватског језика (табела~\ref{table:2}). Ово указује да је кључ за добро угњежђивање реченица претходно обучавање модела за језике који се испитују. 

Дакле, за угњежђивање реченица на српском језику најбољи су модели који су претходно обучени на довољно великом скупу реченица српског језика, али ако се обрађују и реченице на хрватском, модели који су обучавани и на српском и на хрватском имају предност. Са друге стране, модели засновани на \textit{XLM-R} архитектури који су унапред обучени на сто светских језика поново показују најлошије резултате, вероватно због великог шума који разнолики скуп за обучавање производи.


\begin{figure}[ht!]
\centering
\includegraphics[scale=0.25]{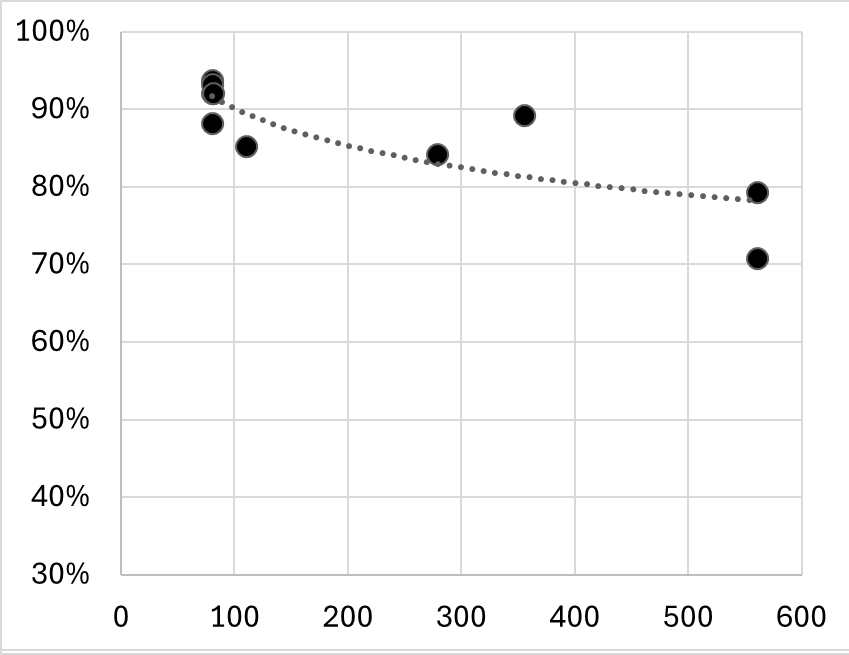}
\includegraphics[scale=0.25]{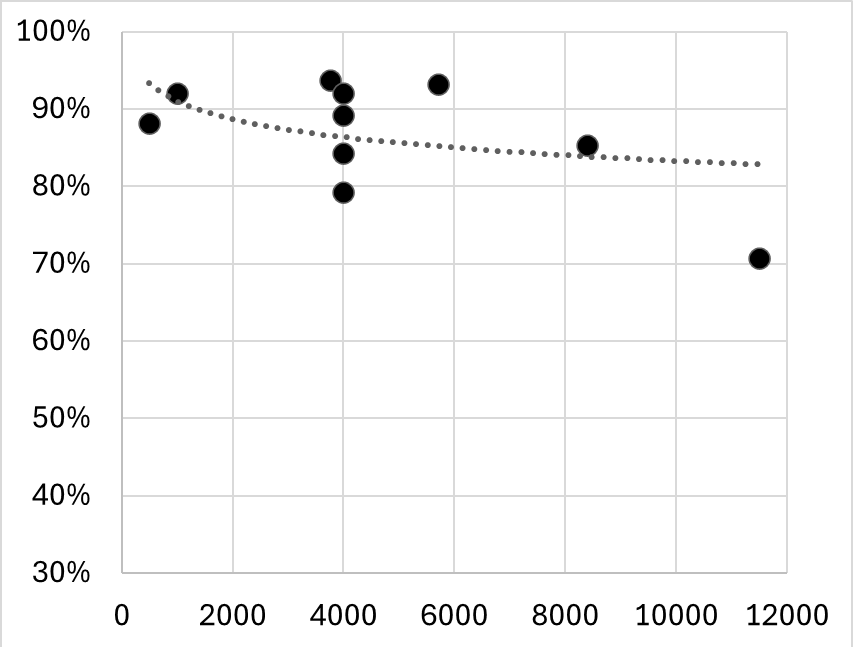}
\caption{Тачност модела на задатку угњежђивања према величини тих модела (лево), као и према величини скупова за обучавање модела (десно). Приказана крива тренда одговара логаритамској функцији.}
\label{fig:slika3}
\end{figure}

На слици~\ref{fig:slika3} приказано је какав је утицај величине модела и скупова за обучавање на перформансе на овом задатку, а линије тренда указују да се са повећањем и модела и скупа за обучавање перформансе смањују. Утицај величине скупа може донекле бити приписан претходно описаном феномену који погађа \textit{XLM-R} архитектуру. Ипак, када је у питању утицај величине модела, постоје додатни индикатори да су мањи модели бољи за овај задатак када је у питању обрада вишејезичних текстова. Тако \textit{jerteh/jerteh-81} надмашује \textit{jerteh/jerteh-355} на задатку утврђивања сличности између реченица на српском и хрватском језику. Разлог може бити то што је мањи модел, услед недостатка у величини, слабије прилагођен српском језику (\textit{model underfit}), али зато има предност приликом генерализације.


\subsection{Низводни задаци}

За разлику од евалуације на узводним задацима, на низводним задацима су резултати које постижу модели много сличнији. На задатку обележавања врстом речи ($T_3$) готово сви модели остварују добре резултате (табела~\ref{table:4}), укључујући и оне засноване на \textit{XLM-R} архитектури. Штавише, \textit{classla/xlm-r-bertic} и \textit{xlm-roberta-large} су два од четири модела који остварују најбоље резултате (друга два модела су \textit{classla/bcms-bertic} и \textit{jerteh/jerteh-355}). 

\begin{figure}[ht!]
\centering
\includegraphics[scale=0.25]{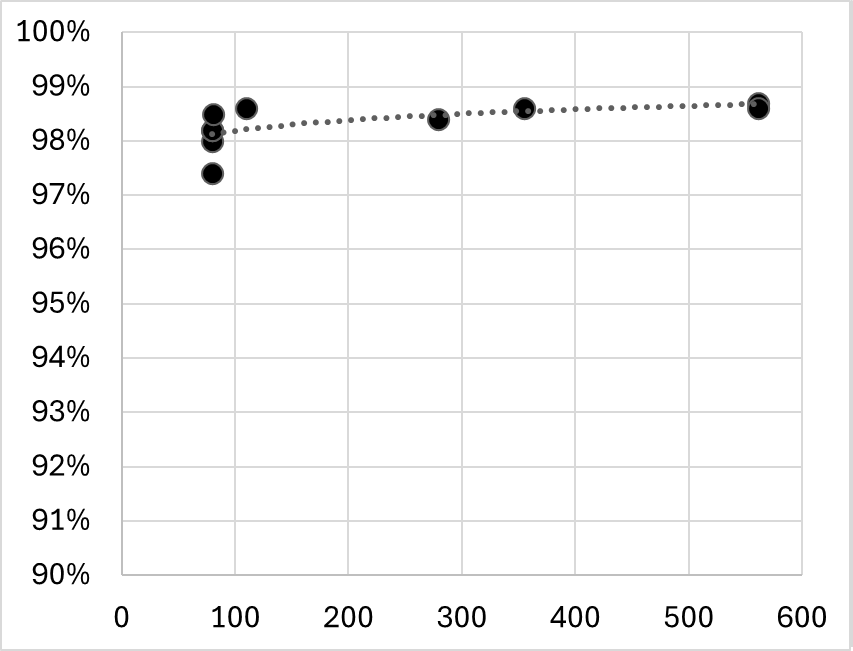}
\includegraphics[scale=0.25]{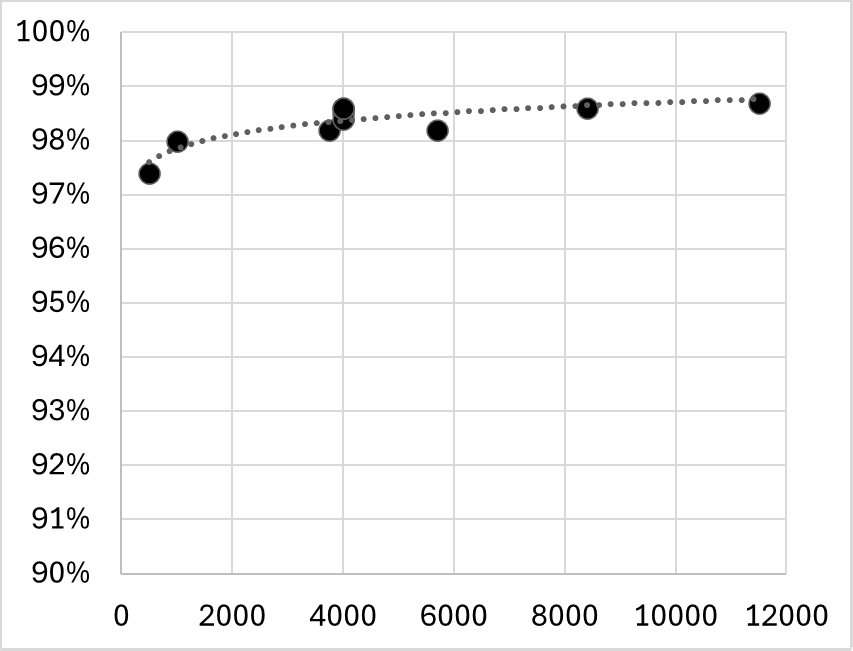}

\caption{Перформансе модела на задатку обележавања врстом речи према величини тих модела (лево), као и према величини скупова за обучавање модела (десно). Приказана крива тренда одговара логаритамској функцији.}
\label{fig:slika4}
\end{figure}

Заједничко за ова четири модела је да су то или највећи модели или модели који су обучавани на највећим скуповима података. Позитивна корелација између перформанси и величине модела, као и између перформанси и величине скупова за обучавање уочава се и слици~\ref{fig:slika4}.


\begin{figure}[ht!]
\centering

\includegraphics[scale=0.25]{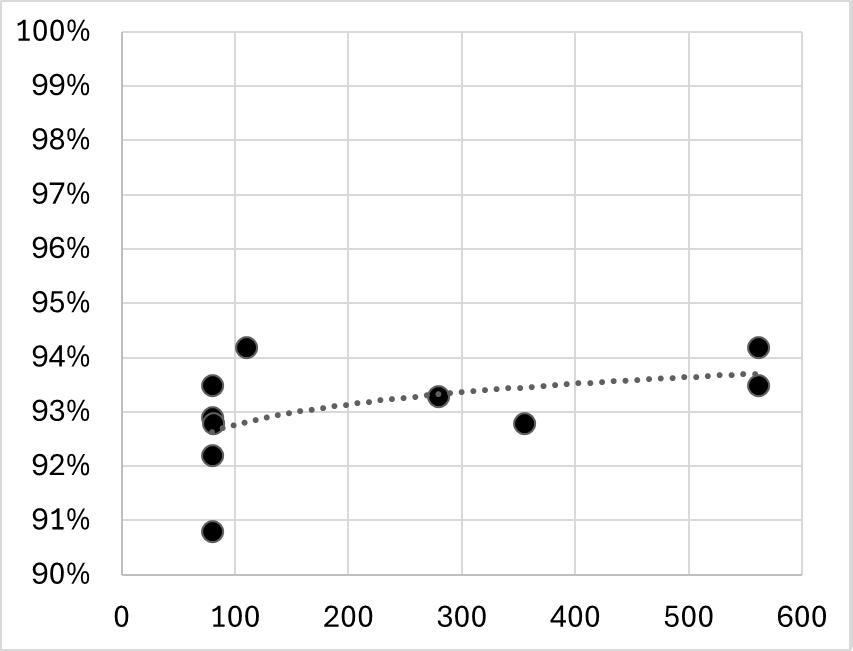}
\includegraphics[scale=0.25]{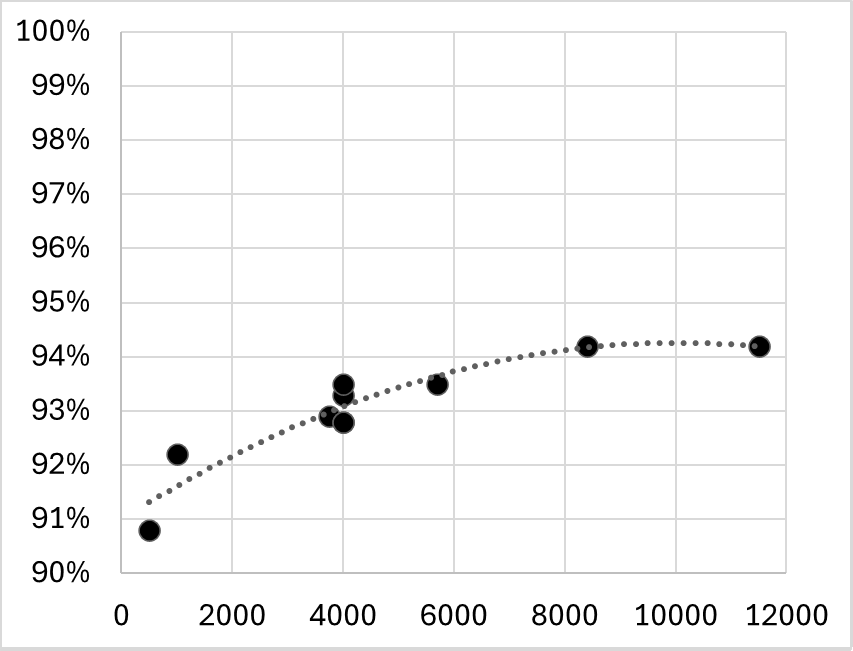}
\caption{Перформансе модела на задатку препознавања именованих ентитета према величини тих модела (лево), као и према величини скупова за обучавање модела (десно). Приказана крива тренда одговара логаритамској функцији.}
\label{fig:slika5}
\end{figure}

Корелација између величине скупа за обучавање још је очигледнија у случају препознавања именованих ентитета (слика~\ref{fig:slika5}). Најбоље резултате на овом задатку (задатку $T_4$) су остварила два модела са највећим скуповима за обучавање, док је поново приметна и успешност \textit{XLM-R} модела, као и код претходног задатака. Величина модела такође показује тек незнатну позитивну корелацију.


\subsection{Закључак}
Када је у питању моделирање маскираног језика, чини се да развој нових модела за српски језик иде у правом правцу. Модел \textit{jerteh/jerteh-355} остварује убедљиво најбоље резултате, бар када је у питању рад са високо-квалитетним текстовима, чак и када су они маскирани токенизаторима других модела (табела~\ref{table:1}). Премда повећање скупа података за обучавање повољно утиче на перформансе модела (слика~\ref{fig:slika2}), не треба занемарити ни квалитет скупа, јер модели \textit{jerteh/jerteh-355} и \textit{jerteh/jerteh-81} надмашују моделе \textit{Andrija/SRoBERTa-F} и \textit{Andrija/SRoBERTa-XL} који су обучени на већим скуповима за обучавање, указујући да веб-корпуси можда нису увек довољни за обучавање добрих модела. Овај закључак је у складу са закључком из једног другог недавног истраживања~\autocite{li2023textbooks}; ипак, ново истраживање би требало да у скуп за евалуацију уврсти и нелитерарне изворе, како би се добила свеобухватнија слика стања.


На задатку израчунавања сличности између реченица, то јест угњежђивања реченица, истакли су се модели као што су \textit{Andrija/SRoBERTa-F} и \textit{Andrija/SRoBERTa-XL}, за којима сасвим мало заостају \textit{Andrija/SRoBERTa-L}, \textit{jerteh/jerteh-81} и \textit{jerteh/jerteh-355}, када су у питању реченице на српском језику (табела \ref{table:3}). Оно што издваја ове моделе је да су они мањи у односу на друге моделе и да су обучавани на већем скупу, па изгледа да је за овај задатак кључна генерализација, дакле, већи скупови података (или можда мањи модели). Такође, када је у питању обрада реченица ширег језичког спектра (нпр. јужнословенски језици) било би потребно да се реченице из комплетног спектра укључе у скуп за обучавање или, још боље, да се речник прилагоди за мапирање ширег спектра токена, па самим тим и за правилну векторизацију ових реченица. Ново истраживање на ову тему би  требало да истражи и модернији начин векторизације реченица, на пример, коришћењем архитектуре трансформера реченица \textit{sentence transformers}~\autocite{reimers2019sentence}.


У случају оба \textit{узводна задатка}, тачност коју остварују модели засновани на \textit{XLM-R} архитектури је знатно нижа у односу на тачност модела заснованих на \textit{RoBERTa} архитектури. У случају првог задатка ($T_1$) то се може објаснити њиховим знатно већим речником токена (па је самим тим и избор одговарајућег токена тежи). Ипак, такво објашњење не би било адекватно и за други задатак ($T_2$). Са друге стране модели засновани на \textit{XLM-R} архитектури су се показали као најбољи (уз малу маргину) на низводним задацима, пре свега на задатку препознавања именованих ентитета ($T_4$). Чини се да је за успешно решавање овог задатка најбоље да се модел током обучавања сусретне са најразличитијим бројем токена, али додатна побољшања доноси и додатно обучавање на српским текстовима. Изгледа да би за  унапређење перформанси тренутно било оптимално дообучавање \textit{XLM-RoBERTa-large} модела коришћењем што већег и квалитетнијег скупа текстова на српском језику. Када је у питању обележавање врстом речи, изгледа да би било који нови модел био адекватан за решавање овог задатка.


\section*{Захвалница}
Најважније скупове података за обучавање модела \textit{GPT2-orao}, \textit{GPT2-vrabac}, \textit{jerteh-81} и \textit{jerteh-355} обезбедило је Друштво за језичке ресурсе и технологије.\footnote{\href{https://jerteh.rs/}{JeRTeh}}

Рачунарске ресурсе за обучавање модела \textit{GPT2-orao} и \textit{GPT2-vrabac} обезбедила je Национална платформа за вештачку интелигенцију Србије.

Рачунарске ресурсе за обучавање модела \textit{jerteh-81} и \textit{jerteh-355} обезбедио је Рударско-геолошки факултет Универзитета у Београду.

Истраживање је спроведено уз подршку Фонда за науку Републике Србије, \#7276, \textit{Text Embeddings – Serbian Language Applications} – \textit{TESLA}.

Хвала!\\

\printbibliography
\label{lastpage}
\end{document}

%% file: settings/settings.tex
\usepackage[english]{babel}
\usepackage[T1,T2A]{fontenc} 
\usepackage[utf8]{inputenc}

\usepackage{csquotes}

\usepackage{amsmath}
\usepackage[authordate,backend=biber,dashed=false]{biblatex-chicago}

\usepackage{geometry}
\usepackage{graphicx}
\graphicspath{ {images/color/} }
\usepackage[dvipsnames]{xcolor}
\usepackage{ragged2e}
\usepackage{array}
\usepackage{changepage}
\usepackage{fancyhdr}
\usepackage{environ}
\usepackage[colorlinks = true,
            linkcolor = Sepia,
            urlcolor  = Mahogany,
            citecolor = BrickRed,
            anchorcolor = Sepia]{hyperref}
\usepackage{indentfirst}
\usepackage{hyperref}
\usepackage{footnote}
\makesavenoteenv{tabular} 

\definecolor{light_gray}{rgb}{0.5, 0.5, 0.5}
\definecolor{dark_gray}{rgb}{0.4, 0.4, 0.4}
\definecolor{darker_gray}{rgb}{0.2, 0.2, 0.2}

\newcommand{\udk}[1]{\noindent\textcolor{dark_gray}{\parbox{\textwidth}{\texttt{УДК #1}}}} 

\renewcommand{\title}[1]{\begin{center}\Large\bfseries\boldmath#1\end{center}}
\renewcommand{\author}[1]{#1 \newline}
\newcommand{\institution}[1]{
	\textcolor{darker_gray}{\textit{\small #1}} \newline
}
\renewcommand{\email}[1]{{\textcolor{light_gray}{\small #1}} \newline}
\newcommand{\orcid}[1]{
	\textcolor{light_gray}{\texttt{\small ORCID: #1}} \newline
}

\newenvironment{main}
		{\begin{tabular}{ l @{~} @{~} !{\vrule width 2pt}| @{~} l }}
		{\end{tabular}}
\NewEnviron{abstractenv}{%
		\parbox[t]{0.6\textwidth}{	
			\begin{adjustwidth}{-1.5cm}{}	
				\begin{abstract}							
					\BODY						
				\end{abstract}
			\end{adjustwidth}
		}
}%
\NewEnviron{authorsenv}{%
  \parbox[t]{0.4\textwidth}{%
      \BODY
   }
}%

\usepackage{comment}

%% file: settings/language_srp.tex
%
%
\renewcommand{\thesection}{\arabic{section}.} 
\renewcommand{\thesubsection}{\arabic{section}.\arabic{subsection}} 

%
%
\renewcommand{\keywords}[1]{\textbf{\noindent КЉУЧНЕ РЕЧИ:} \raggedright #1}
\renewcommand{\abstractname}{\noindent САЖЕТАК:} 
\renewcommand\refname{Литература}
\renewcommand\tablename{Табела}
\renewcommand\figurename{Слика}
\newcommand{\arrivaldate}[1]{{\textbf{РАД ПРИМЉЕН:}}{{\hfill #1}} \\}
\newcommand{\acceptdate}[1]{{\textbf{РАД ПРИХВАЋЕН:}}{{\hfill #1}}}

\newcommand{\dates}[3]{%
	\vspace{#1cm}
	\noindent\parbox[b]{0.57\textwidth}{%
		\arrivaldate{#2} 
		\acceptdate{#3} 
	}	
}%